# HSADML: Hyper-Sphere Angular Deep Metric based Learning for Brain Tumor Classification

Aman Verma[1] and Vibhav Prakash Singh[2]

[1] Dept. of Electronics and Communications, National Institute of Technology Raipur
`aman.verma.nitrr@gmail.com`
[2] Department of Computer Science & Engineering, Motilal Nehru National Institute of Technology Allahabad, Prayagraj, 211004, India
`vibhav@mnnit.ac.in`

**Abstract.** Brain Tumors are abnormal mass of clustered cells penetrating regions of brain. Their timely identification and classification help doctors to provide appropriate treatment. However, Classification of Brain Tumors is quite intricate because of high-intra class similarity and low-inter class variability. Due to morphological similarity amongst various MRI-Slices of different classes the challenge deepens more. This all leads to hampering generalizability of classification models. To this end, this paper proposes HSADML, a novel framework which enables deep metric learning (DML) using SphereFace Loss. SphereFace loss embeds the features into a hyperspheric-manifold and then imposes margin on the embeddings to enhance differentiability between the classes. With utilization of SphereFace loss based deep metric learning it is ensured that samples from class clustered together while the different ones are pushed apart. Results reflects the prominence in the approach, the proposed framework achieved state-of-the-art 98.69% validation accuracy using k-NN (*k=1*) and this is significantly higher than normal SoftMax Loss training which though obtains 98.47% validation accuracy but that too with limited inter-class separability and intra-class closeness. Experimental analysis done over various classifiers and loss function settings suggests potential in the approach.

## 1 Introduction

Brain tumors are malicious accumulation of anomalously growing cells in the brain parenchyma and the regions in the vicinity. Early-stage diagnosis of brain tumors helps medical experts to take timely medication. Along with diagnosis it is of utmost importance that the correct class of the tumor is also identified. The task of brain tumor classification can be categorized as a three-class detection problem wherein the classes are Meningioma, Glioma and Pituitary Tumor. Computer Aided Diagnosis (CAD) based approaches have become prevalent in classification and more recently machine intelligence-based solutions are being employed to augment CAD frameworks. To be specific, deep learning-based methods have established dominance in the domain owing to their robust performances [1]. These approaches utilize deep neural network architectures such as deep CNN to capture complex nonlinear decision boundaries. Multiple attempts have been done to enhance classification performances, in [2] authors used a 3-stage approach to classify the grade of the tumors, first the brain-tumors were segmented, then from the segmented images data augmentation was done after which a CNN was employed for feature extraction cum classification. Transfer Learning approaches involving ImageNet [3] pre-trained models have been actively employed to achieve state-of-the-art performance [4]. To overcome overfitting GAN based brain MRI data augmentation was done in [5] while approaches involving Capsule Networks have been recently proposed [6]. BrainMRNet [21] has been recently introduced, it uses Attention Mechanism [22] to bring robustness in features. Even though there has been significant progress so far, there is a wide scope for improvements. Firstly, there is need to improve generalizability of the models and secondly extraction of more prominent features is also essential. High inter-class similarity along with high intra-class variability degrades the performance. Foremost cause of the same is that the visual morphology and view-of-capture of the MRI-slices among the classes is quite similar. Figure 1 depicts structural similarity between the three brain tumor types. Moreover, the major tumor-type differentiating information remains concealed with limited spatial space. Thus, this research attempts to overcome above mentioned problems and tries to improve feature discrimination to enhance performance and generalization of brain tumor classification models.

Deep metric learning has been a natural choice in multiple domains [7,8,20] to enforce inter-class separability and intra-class similarity. Usually, SoftMax based classifica-tions are performed but it is to be noted that SoftMax loss maximizes class distances so as to have features before the last fully connected layer linearly separable as it is a softened version of the max operator. This enforces low inter-class separation which causes minimal intra-class similarity. Deep metric learning (DML) on the other hand metricizes the distance between training examples which makes examples belonging to the same class being clustered while the different ones moving towards respective class centers leading to inter-class separation. DML is applied via training deep learning models with distance aware losses. Triplet Loss [9] has been one of the most widely used DML loss, for any example(anchor) it samples a positive(same-class) and negative(different-class) example, computes distance between embeddings of anchor and positive and anchor and negative, then it imposes a Euclidean margin between them to improve inter-class distances and intra-class closeness. Though it has been successful, selection of inadequate margin and sampling wrongly may lead to unstable training.

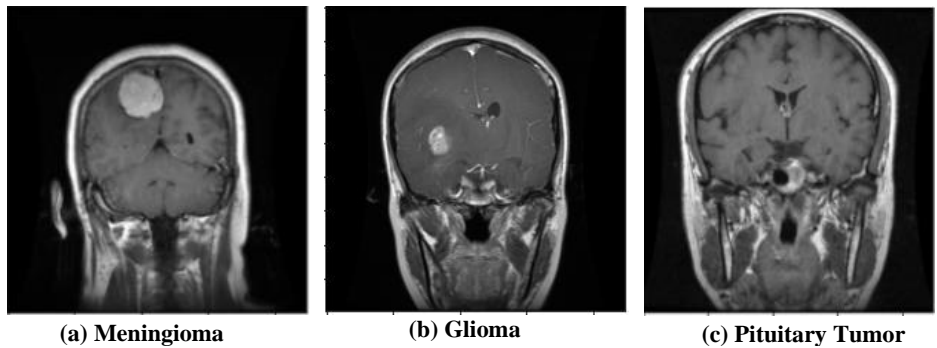

(a) Meningioma    (b) Glioma    (c) Pituitary Tumor

**Figure 1: MRI Scan slices from each of the Brain Tumor types**

Concepts of adaptive margins and hard-negative mining [10] have been introduced to alleviate the problems in prior but issue related sampling sustains. Other loss functions such as Center-Loss [11] supervise SoftMax loss through another term that minimizes intra-class variance, but in such losses SoftMax loss empirically dominates. An advancement of triplet loss has been the Quadruplet Loss [12] which samples another negative with a target to set minimum inter-class distance greater than maximum intra-class distance, but again negative sampling remains an issue of concern. Angular Loss [13] improvised over the previous works by enforcing angular margin and hence utilizing gradients from all negative, positive and anchor, this was lacking in Euclidean margin-based triplet loss strategies but negative mining must be done here also.

To alleviate shortcomings this paper, propose HSADML framework – a hyper-sphere manifold metric-based learning for brain tumor classification. The proposed approach utilizes SphereFace Loss [14] which surpasses the tedious nature of triplet loss-based sampling and enforces angular separation using angular margin. To the best of knowledge this is the first work on brain tumour classification to utilize DML. Following is the summary of key contributions: -

- HSADML Framework to enhance generalizability of brain tumor classification models.
- HSADML framework uses SphereFace loss that alleviates the issue of triplet sampling but at the same time develops discriminative representation using angular margin based Hyper-Spherical DML.
- The proposed approach achieves state-of-the-art results on benchmark dataset. Extensive experimentation done to verify the methodology suggests the same.

This paper has been organized in four sections The current sectioned gave an introduction and literature review. Proposed framework is explained in next section, while in the third section – Experiments, Results and Discussions, experimental analysis bas been done. Finally, the paper is concluded in Section 4 with future research dimensions

## 2  Methodology

HSADML Framework has two major components first being the SphereFace Loss Function and the second being the backbone network as diagrammatically illustrated in the Figure 2. In this study, MobileNet [15] has been utilized as the backbone network with reason being lightweight nature of the same. All the components of HSADML framework have been explained in subsequent subsections.

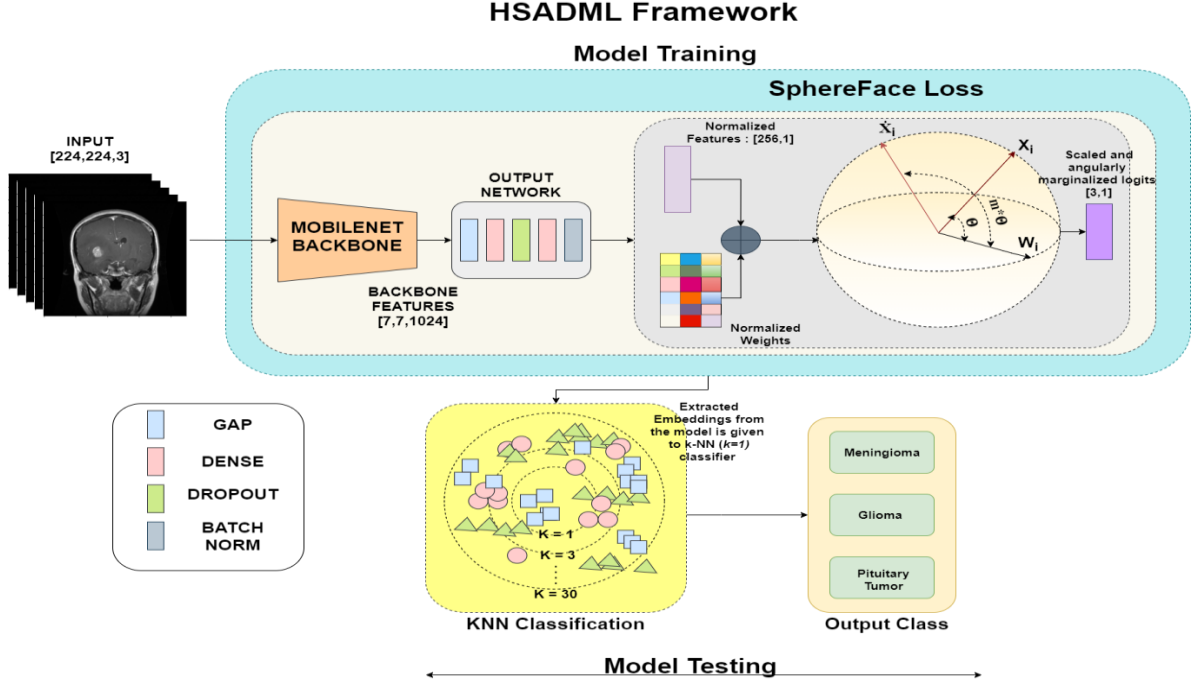

**Figure 2:** HSADML framework has two phases - the training phase and the testing phase. In training phase, the deep learning model is trained via angular metric learning using SphereFace Loss. While in testing phase k-NN (*k=1*) based testing is done.

### 2.1 SphereFace Loss

It is visually challenging to identify the type of brain-tumor mainly because of two reasons – 1) The affected region being quite small, at large scale bears resemblance for all the three class. 2) Multi-view data appends morphological similarity which further enforces inter-class compactness and intra-class discrepancy. SoftMax Loss function on the other hand when employed in classifiers generates linearly separable decision boundaries using the features of the last dense layer. Mathematically, the SoftMax Loss function is defined as follows:

$$\mathcal{L}_{Softmax} = -\frac{1}{N}\sum_{i=1}^{N} \log \frac{e^{(W_{yi}^T x_i)}}{e^{(W_{yi}^T x_i)} + \sum_{j \neq i} e^{(W_{yj}^T x_i)}} \quad (1)$$

Here, $x_i \in \mathbb{R}^d$ with d being the dimension of feature embeddings. Let there be n classes - $[y_1, y_2, \cdots, y_n]$ and the batch-size is N, $W \in \mathbb{R}^{d*N}$ represents the weights while $W_{yi} \in \mathbb{R}^{1Xd}$ is the weight column corresponding to class $y_i$. For the sake of brevity, the bias term hasn't been included. For SoftMax Loss linear-vector space is considered for classification where the class boundaries are bounded within a certain defined region, this restricts the generalizability of the model. Whereas, if the classification is done in angular some specific sector gets assigned to particular class which makes the region of existence for that class unbounded within the sector. HSADML framework on the other hand utilizes SphereFace Loss Function to facilitate angular metric learning. SphereFace Loss projects the feature embeddings onto a hyper-sphere manifold by normalizing both weights and feature embeddings. Then, it metricizes the angular distance between class centers0 and the feature embeddings. By imposing an angular margin, it makes training harder but ensures intra-class compactness and inter-class variance. $W_{yj}^T x_i$ defines the dot product between $W_{y_j}$ and $x_i$, thus SphereFace Loss formulates following Euclidean to Angular Space Transformation - $\|W_{y_j}\| \|x_i\| \cos \theta_{y_j i}$. Here, $\theta_{y_j i}$ denotes the angle between $W_{y_j}$ and $x_i$. The class center and the feature embeddings are $L_2$-normalized which makes the transformation dependent only on $\theta_{y_j i}$. The normalization step projects the embeddings on a $d$ dimensional hypersphere. Finally, angular margin $m$ is introduced to make training tougher but this results in formation intra-class clustered and inter-class separated embeddings. Formally, SphereFace Loss can be defined as follows:

$$\mathcal{L}_{SpF} = -\frac{1}{N}\sum_{i=1}^{N} \log \frac{e^{\|s\|\Psi(\theta_{y_i i})}}{e^{\|s\|\Psi(\theta_{y_i i})} + \sum_{j \ne i} e^{e^{\|s\|\cos(\theta_{y_j i})}}} \quad (2)$$

Here,

$$\Psi(\theta_{y_i i}) = (-1)^k \cos(m\theta_{y_i i}) - 2k \quad (3)$$

$\|s\|$ is a scaling factor, $k \in [0, m-1]$, $\theta_{y_j i} \in [\frac{k\pi}{m}, \frac{(k+1)\pi}{m}]$ and $m \geq 4$. Function $\Psi(.)$ has been utilized to mitigate instability-intricacies during CNN training.

Angular margin makes the training more challenging; by increasing the angular distance of feature embedding with its respective ground-truth class center by a factor of *m*. This results in model learning more robust embeddings that stand-out on magnifying inter-class separation while minimizing intra-class separation. Due to classification now being done angularly on hyper-spherical manifold generalization of the model is further enhanced. Furthermore, the loss function does not inculcate triplet/quadruplet sample mining.

## 2.2 Backbone Architecture

SphereFace Loss requires feature-rich embeddings to operate smoothly over the hyper-spherical deep metric learning. Thus, HSADML framework incorporates the use of ImageNet pretrained model of MobileNet. Specifically, MobileNet architecture has been chosen for the base network because on low-computational burdens that it lifts. The depthwise separable and point-wise convolutions drastically reduce the number of parameters in MobileNet while supports good performance. Accredited to the same, inference time of the MobileNet model is quite low. This all ensures that there is scalability in approach for practical deployments in medical diagnosis.

The enriched representation extracted out of the MobileNet model are fed to Output Network wherein the feature-maps are firstly average pooled globally (GAP Layer), then the pooled embeddings are passed through a series of dense layers and dropout layers. Finally, after the final dense layer which is *He Instantiated* [16] and linearly activated, a BatchNormalization layer [17] is applied which helps in making the training stable. These are the final embeddings which are passed onto the SphereFace Loss function. In this case dimensionality of the final embeddings was chosen to be (256,1).

### 2.3    Classification and Testing

After the model has been trained under SphereFace Loss, embeddings from the final dense layer are extracted. Then using cosine distance as distance metric k-NN (k=1) algorithm is employed so as to test the model. The extracted embeddings are $L_2$ normalized, and instead of directly applying cosine-distance metric, Euclidean metric over normalized embeddings was utilized. The normalization step ensures that there is proportionality between the respective Euclidean and Cosine distance. Mathematically, let there be two normalized embeddings $x_{i1}$ and $x_{i2}$, then:

$$\text{cosdist}(x_{i1}, x_{i2}) = 1 - x_{i1}^T x_{i2} \qquad (4)$$

Similarly,

$$\text{eucdist}(x_{i1}, x_{i2}) = [2 * (1 - x_{i1}^T x_{i2})]^{\frac{1}{2}} \qquad (5)$$

This makes,

$$\text{eucdist}(x_{i1}, x_{i2}) \equiv \text{cosdist}(x_{i1}, x_{i2}) \qquad (6)$$

Here, cosdist(.) represents the cosine distance and eucdist(.) the Euclidean distance.

k-NN with first nearest neighbor being a very basic classifier, HSADML based extracted embeddings are also classified using more sophisticated classifiers. Specifically, SVM with Gaussian and Polynomial kernel, Random Forest and k-NN (with best possible nearest neighbor combination considering up to 30 neighbors) has also been employed to facilitate robust classification of the extracted feature embeddings. For training all the classifiers, embeddings are extracted from both training and testing set which after which the classifier model is trained with the embeddings from training set and for inference-evaluation testing set extracted embeddings are used.

### 2.4    Implementation and Network Training

The input to the model were 3-channel stacked MRI images with image dimensions being (224,224,3). Prior to inputting the images rescaling of pixel-intensities are rescaled to range [0,1]. Data Augmentation strategies involving use of random rotation, random brightness increment-decrement, random flipping and zooming-in are applied over the inputs. The augmented data is given to the MobileNet which produces 1024 feature maps each having height and width as (7,7) respectively. In the Output Network, the first dense layer transforms the feature dimension from 1024 to 256, then a Dropout layer [20] is applied with rate of 0.2. For the SphereFace Loss function, experimentation on margin is done and the empirically found optimal margin i.e. m=5 is considered while the scaling factor was set to 30. Stabilization in model performance was seen with the application of BatchNormalization after the final dense layer. The model was trained with Adam Optimizer and for 275 epochs with a custom learning rate schedule.

$$lr = \begin{cases} 1e-4 & if\ epoch < 125 \\ 1e-5 & if\ 125 \leq epoch < 175 \\ 1e-6 & otherwise \end{cases}$$

This schedule was designed so as to slower the learning rate when convergence has been attained. For the case of SphereFace Loss function the model converged started from about 100 epochs but loss kept optimizing till the end and consequently the embeddings kept becoming better. Loss optimization for the proposed HSADML model has been illustrated in the Figure 3. The following curve suggests smooth convergence.

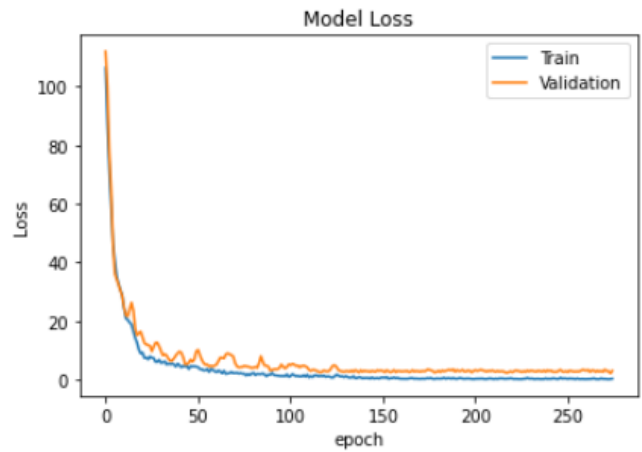

**Figure 3: Loss optimization during training of HSADML framework**

## 3     Experiments, Results and Discussions

### 3.1    Dataset and Experimental Protocol

All the experimentation on the HSADML framework has been done using the Benchmark Dataset of Figshare [18]. It contains 3064 T1 weighted FLAIR images of Brain MRI with 708, 1426 and 930 belonging respectively to Meningioma, Glioma and Pituitary tumor class. As suggested in [21] all the results were reported on 70-30% training-testing data split, but to elucidate the robustness in the approach results over 30-70% training-testing data-split is also presented. These experiments involved usage of hold-out evaluation strategy wherein the testing data of 70-30% data-split became the training data of 30-70% data split; vice-versa is also true. Along with analyzing the performance of model over different data-settings, experimentation over the SphereFace Loss and different classifiers have been done but these experiments involved only the use of 70-30 data-split.

## 3.2 Performance Metrics

Models involved in this study are compared both graphically and numerically. For numeric performance evaluation following metrics have been used:

- **Accuracy:** Both intra-class and overall accuracy of the models have been computed. Let True Positives, False Negatives, True Negatives and False positive be represented by TP, FN, TN, FP. Then

$$Accuracy = \frac{TP + TN}{TP + FP + TN + FN}$$

- **F1-Score:** F1-Score represents the harmonic mean computed over both Precision ($\frac{TP}{TP+FP}$) and Recall ($\frac{TP}{TP+FN}$). It is a good metric for data-imbalance and also finds its significance for the cases where FN and FP have different prominence. It has also been computed over per-class and overall. Mathematically,

$$\frac{2}{F1} = \frac{1}{Precision} + \frac{1}{Recall}$$

- **Matthews Correlation Coefficient (MCC):** MCC takes into account entire confusion matrix and is a good evaluation measure for imbalanced-data.

$$MCC = \frac{TP*TN - FP*FN}{[(TP+FP)*(TP+FN)*(TN+FP)*(TN+FN)]^{\frac{1}{2}}}$$

- **Area Under Curve (AUC):** AUC value is the area under the ROC curve. Higher the AUC better the True Positive Rate and Lower the False Positive Rate.

## 3.3 Experimental Analysis

This section discusses on the results of various experiments and analysis over them. Firstly, the performance evaluation of the HSADML Framework over 70-30% and 30-70% training-testing data split has been done. The results have been tabulated in Table 1. For the sake of simplicity, we refer each class by an abbreviation i.e., Meningioma (Me), Glioma (Gl) and Pituitary Tumor (Pt). The proposed framework at 70-30% data setting achieves overall high accuracy of 98.69% while maintain a decent

Table 1 Performance Analysis of HSADML Framework for Different Data Split

| Training Dataset (in %) | Testing Dataset (in %) | Accuracy (in %) | | | | F1-Score | | | | | MCC |
|---|---|---|---|---|---|---|---|---|---|---|---|
| | | Me | Gl | Pt | Avg. | Me | Gl | Pt | Macro Avg. | Micro Avg. | |
| 70 | 30 | 95.85 | 99.76 | 99.28 | 98.69 | 0.9719 | 0.9964 | 0.9841 | 0.9841 | 0.9869 | 0.9796 |
| 30 | 70 | 89.00 | 97.01 | 97.99 | 95.47 | 0.9104 | 0.9667 | 0.9687 | 0.9486 | 0.9547 | 0.9289 |

Table 2 Comparison of HSADML framework generated embeddings using different machine learning classifiers

| Classifier | Accuracy (in %) | | | | F1-Score | | | | | MCC |
|---|---|---|---|---|---|---|---|---|---|---|
| | Me | Gl | Pt | Avg. | Me | Gl | Pt | Macro Avg. | Micro Avg. | |
| k-NN *(k=1)* | 95.85 | 99.76 | 99.28 | 98.69 | 0.9719 | 0.9964 | 0.9841 | 0.9841 | 0.9869 | 0.9796 |
| Random Forest | 94.93 | 99.76 | 99.28 | 98.47 | 0.9671 | 0.9941 | 0.9846 | 0.9817 | 0.9847 | 0.9762 |
| SVM (Gaussian Kernel) | 95.85 | 99.76 | 99.28 | 98.69 | 0.9719 | 0.9964 | 0.9841 | 0.9841 | 0.9869 | 0.9796 |
| SVM (Polynomial Kernel) | 96.77 | 98.81 | 98.57 | 98.26 | 0.9630 | 0.9940 | 0.9805 | 0.9792 | 0.9826 | 0.9729 |

MCC score of 0.9762. The average accuracy of the model is significantly high for the classes Gl and Pt but little less for Me which turns out to be the most challenging class. F1-Score on class as well as on overall basis suggest significant performance. In the 30-70% data split, training data is quite limited in amount. Although in this scarce training data experiment, performance is retained. Average accuracy achieved in this

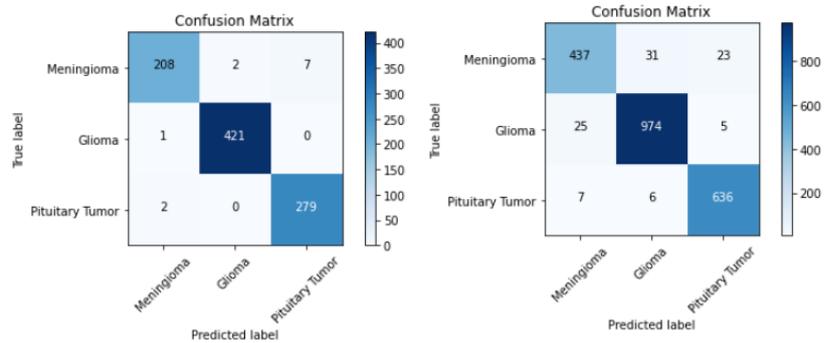

(a) 70-30 % Data Split's Confusion Matrix    (b) 30-70 % Data Split's Confusion Matrix

**Figure 4: Confusion Matrices of HSADML Framework**

Table 3 Comparison of HSADML framework with Various Loss Functions

| Loss | Accuracy (in %) | | | | F1-Score | | | | | MCC |
|---|---|---|---|---|---|---|---|---|---|---|
| | Me | Gl | Pt | Avg. | Me | Gl | Pt | Macro Avg. | Micro Avg. | |
| SoftMax Loss | 95.39 | 99.05 | 100.00 | 98.47 | 0.9695 | 0.9940 | 0.9825 | 0.9820 | 0.9847 | 0.9763 |
| Modified SoftMax Loss | 94.93 | 99.76 | 99.28 | 98.47 | 0.9671 | 0.9941 | 0.9846 | 0.9817 | 0.9847 | 0.9762 |
| Triplet Loss | 96.31 | 99.28 | 99.28 | 98.58 | 0.9698 | 0.9952 | 0.9891 | 0.9830 | 0.9858 | 0.9797 |
| SphereFace Loss (Proposed) | 95.85 | 99.76 | 99.28 | 98.69 | 0.9719 | 0.9964 | 0.9841 | 0.9841 | 0.9869 | 0.9796 |

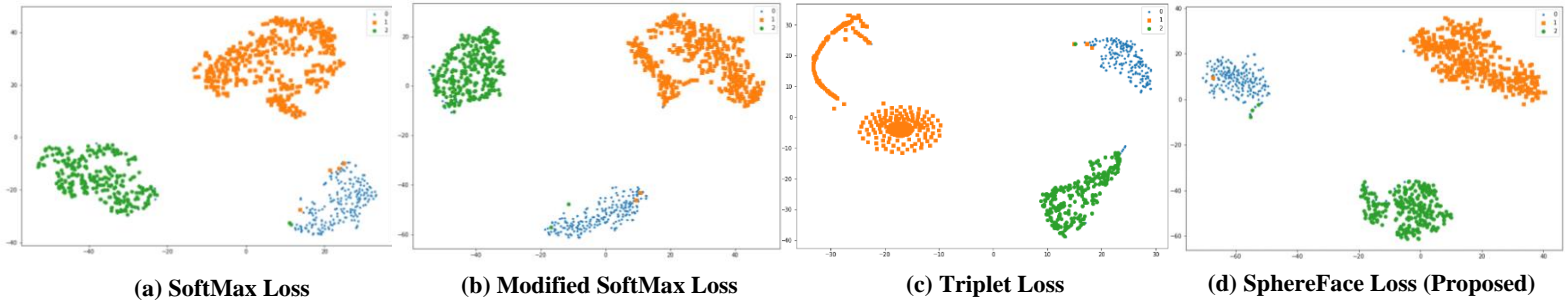

(a) SoftMax Loss    (b) Modified SoftMax Loss    (c) Triplet Loss    (d) SphereFace Loss (Proposed)

**Figure 5: t-SNE comparison of HSADML framework's extracted embeddings with that of ones extracted with other loss functions. Blue color represents Meningioma Class, Orange Glioma and Green Pituitary Tumor**

experiment is 95.47% while the MCC is 0.9289. Figure 4 shows the Confusion Matrix of both the models trained under both the data splits. From the matrix high-end performance over the Gl and Pt class can be noted but there remains a gap of improvement for Me class. More specifically, recall of the models for the Me class is comparatively lesser than the precision. It can be inferred that the base-network utilized in the study is MobileNet, and hence when some more advanced backbone network shall be used, further performance gain is expected.

To further explore capabilities of the proposed model, advanced classifiers have been tested with 70-30% Data-Split's model extracted embeddings. Specifically, SVM with Gaussian and Polynomial kernel, Random Forest and k-NN (k=1) (k=1 was the best possible nearest neighbor combination with considering up to 30 neighbors) has also been employed. The results of the same has been tabulated in Table 2. For both k-NN and SVM with gaussian kernel, the model gave the same classification. Random Forest

Table 4 AUC based comparison of HSADML framework with Various Loss Functions

| Loss | Me | Gl | Pt | Avg. |
|---|---|---|---|---|
| SoftMax Loss | 0.9748 | 0.9942 | 0.9921 | 0.9870 |
| Modified SoftMax Loss | 0.9741 | 0.9964 | 0.9878 | 0.9861 |
| Triplet Loss | 0.9780 | 0.9954 | 0.9909 | 0.9880 |
| SphereFace Loss (Proposed) | 0.9771 | 0.9968 | 0.9909 | 0.9882 |

Table 5 Ablation Study over Angular Margin

| Margin | Accuracy (in %) | | | | F1-Score | | | | | MCC |
|---|---|---|---|---|---|---|---|---|---|---|
| | Me | Gl | Pt | Avg. | Me | Gl | Pt | Macro Avg. | Micro Avg. | |
| $m = 4$ | 95.85 | 99.05 | 99.64 | 98.47 | 0.9674 | 0.9917 | 0.9876 | 0.9822 | 0.9847 | 0.9762 |
| $m = 6$ | 94.00 | 99.52 | 99.64 | 98.26 | 0.9622 | 0.9952 | 0.9790 | 0.9788 | 0.9826 | 0.9730 |
| $m = 5$ | 95.85 | 99.76 | 99.28 | 98.69 | 0.9719 | 0.9964 | 0.9841 | 0.9841 | 0.9869 | 0.9796 |

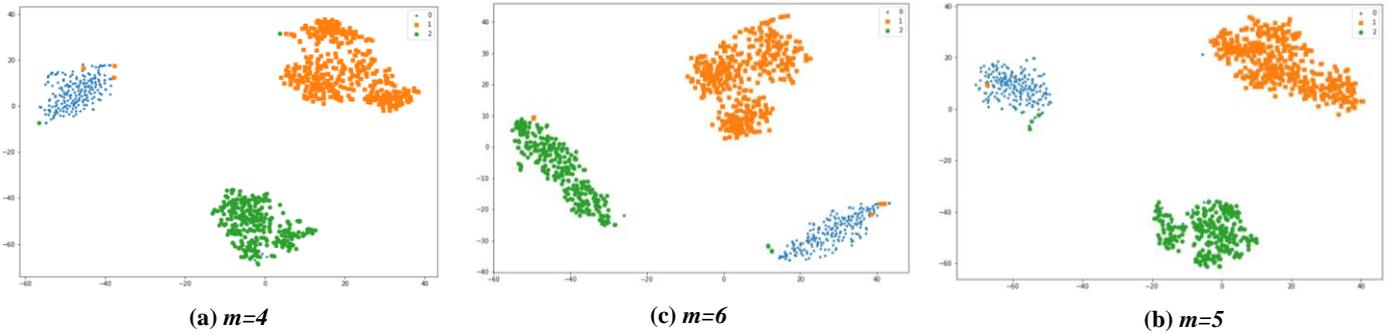

(a) *m=4*      (c) *m=6*      (b) *m=5*

**Figure 6: t-SNE based graphical ablation over different angular margins. HSADML framework uses *m=5* as the angular margin. Blue color represents Meningioma Class, Orange Glioma and Green Pituitary Tumor**

classifier using 400 estimators each utilizing 32 features showed a sustainable performance while SVM with polynomial kernel of order 256 too obtained decent performance. The k-NN achieving the highest performance scores illustrates feature-level discriminability amongst various classes. SphereFace loss enjoyed greater stability in training than the SoftMax Loss model. It is clear from the Table 3 that the proposed HSADML Framework facilitates comparatively higher performance than the other models. The obtained result further justifies on the use of angular margin. In terms of average accuracy, the proposed approach is superior than its counterparts of SoftMax Loss and Modified SoftMax Loss by 0.22 % and 0.33% respectively. In similar manner, the performance of SphereFace Loss is also better than the Triplet Metric Loss, although the model trained with triplet loss better performance for the Meningioma class. In Table 4, for each class Area Under Curve (AUC) value calculated via ROC curves has been tabulated. It is worthwhile mentioning that AUC of SphereFace Loss model is the highest on overall basis as well as for class Gl, while the same is comparable to the best for Me and Pt classes. Further, potential in the approach can be concluded Figure 5 which shows t-SNE Comparison amongst different Loss functions; from the figure, intra-class compactness and inter-class separability can be well spotted for the SphereFace Loss. For the case triplet-loss, intra-class variation appears to be higher while accredited to angular metric, intra-class clustering is observed for the SphereFace

Loss. Hence, it can also be concluded that with training in hyper-spherical domain but without any angular margin, intra-class separation and inter-class discrepancy are not mitigated in comparison to SphereFace Loss.

For understanding the behavior of the angular margin, an ablation amongst different angular margins have been conducted. In this analysis $m = \{4,5,6\}$ has been considered. Empirically, the proposed model with *m=5* emerges out to be better in balancing the trade-off between hard-training and margin-based separation, thereby achieving the best performance. Quantitative results are presented in Table 5 while Figure 6 graph-ically compares the SphereFace Loss variants using t-SNE plot. From the figure a challenging competition lies between *m=4* and *m=5* models, though *m=5* model attains higher inter-class separation. AUC based comparison of models trained with different angular margins has been shown in Table 6. The attained AUC value of the *m=5* curves are higher than the other variants. This trend can be identified specifically for the Gl and Me class, while for the Pt class the model trained with *m=4* attained slightly better result. From the following analysis over angular margins, it can be concluded that angular margin makes the training challenging which results in robust performance, but if the margin is not adequately tuned, then model will converge sub-optimally.

Table 6 AUC based comparison between models trained under different angular margins

| Margin | Me | Gl | Pt | Avg. |
|---|---|---|---|---|
| *m = 4* | 0.9757 | 0.9922 | 0.9935 | 0.9871 |
| *m = 6* | 0.9679 | 0.9956 | 0.9896 | 0.9843 |
| *m = 5* | 0.9771 | 0.9968 | 0.9909 | 0.9881 |

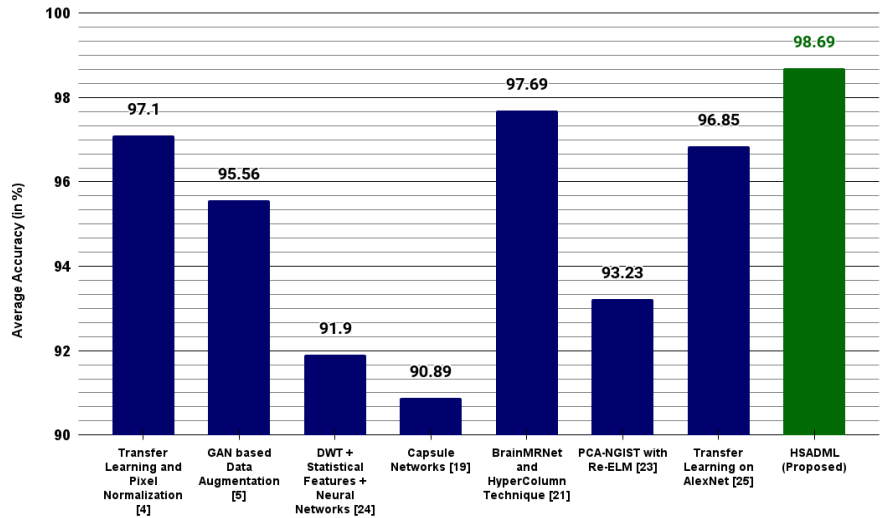

**Figure 7: Comparison of HSADML framework with state-of-the-art methods**

Next part of the analysis is dedicated to comparison of the HSADML framework with the other state-of-the-art approaches. A bar graph depicting comparative analysis of average accuracy between approaches has been plotted in Figure 7. Since, most of the previous works [4,21] consider evaluation on 70-30%, training-testing data-split, we compare our results using the same protocol. It is evident from the same that proposed HSADML framework outperforms the other approaches while obtaining encouraging results. It is to be pointed out that HSADML framework attains better performance for each class, this makes it highly generalized. Thus, basing learning on angular distances in a hyper-spherical manifold can help in optimizing intra-class compactness while increasing inter-class distances. This brings significant improvement in the performance.

## 4      Conclusion and Future Aspect

This paper proposed HSADML framework for brain tumor classification. The proposed approach introduced use of deep angular metric learning using SphereFace Loss for facilitating generalization and robustness in identification. The loss was instrumental in increasing intra-class separability and reducing in intra-class variability; this resulted in achieving significant performance gains. This research didn't emphasize much over the backbone network thus is future aspects of the work there is gap for introduction of some attention-based domain-specific network. Another aspect of the research shall explore on in-depth validation of approaches which has been hampered due to a single-dataset. Nevertheless, high-end performance and lightweight backbone of HSADML framework motivates its scalability and extends application for other modalities as well.

## References

1. Justin S. Paul, Andrew J. Plassard, Bennett A. Landman, and Daniel Fabbri, 2017. Deep learning for brain tumor classification. In *Medical Imaging 2017: Biomedical Applications in Molecular, Structural, and Functional Imaging* (Vol. 10137, p. 1013710). International Society for Optics and Photonics. DOI: https://doi.org/10.117/12.2254195.
2. Muhammad Sajjad, Salman Khan, Khan Muhammad, Wanqing Wu, Amin Ullah, and Sung Wook Baik, 2019. Multi-grade brain tumor classification using deep CNN with extensive data augmentation. *Journal of computational science*, *30*, 174-182. DOI: https://doi.org/10.1016/j.jocs.2018.12.003.
3. Jia Deng, Wei Dong, Richard Socher, Li-Jia Li, Kai Li, and Li Fei-Fei, 2019. A large-scale hierarchical image database. In *2009 IEEE conference on computer vision and pattern recognition* (pp. 248-255). Ieee. DOI: https://doi.org/ 10.1109/CVPR.2009.5206848.
4. S. Deepak, and P. M. Ameer, 2019. Brain tumor classification using deep CNN features via transfer learning. *Computers in biology and medicine*, *111*, 103345. DOI: https://doi.org/10.1016/j.compbiomed.2019.103345.

5. Navid Ghassemi, Afshin Shoeibi, and Modjtaba Rouhani, 2020. Deep neural network with generative adversarial networks pre-training for brain tumor classification based on MR images. *Biomedical Signal Processing and Control*, 57, 101678. DOI: https://doi.org/10.1016/j.bspc.2019.101678.
6. Parnian Afshar, Konstantinos N. Plataniotis, and Arash Mohammadi, 2020. BoostCaps: A Boosted Capsule Network for Brain Tumor Classification. In *2020 42nd Annual International Conference of the IEEE Engineering in Medicine & Biology Society (EMBC)* (pp. 1075-1079). IEEE. DOI: https://doi.org/ 10.1109/EMBC44109.2020.9175922.
7. Krati Gupta, Daksh Thapar, Arnav Bhavsar, and Anil K. Sao, 2019. Deep metric learning for identification of mitotic patterns of HEp-2 cell images. In *Proceedings of the IEEE/CVF Conference on Computer Vision and Pattern Recognition Workshops* (pp. 0-0). DOI: https://doi.org/10.1109/CVPRW.2019.00141
8. Dong Yi, Zhen Lei, Shengcai Liao, and Stan Z. Li, 2014. Deep metric learning for person re-identification. In *2014 22nd International Conference on Pattern Recognition* (pp. 34-39). IEEE. DOI: https://doi.org/ 10.1109/ICPR.2014.16.
9. Florian Schroff, Dmitry Kalenichenko, and James Philbin., 2015. Facenet: A unified embedding for face recognition and clustering. In *Proceedings of the IEEE conference on computer vision and pattern recognition* (pp. 815-823). DOI: 10.1109/CVPR.2015.7298682.D
10. Daksh Thapar, Gaurav Jaswal, Aditya Nigam, and Vivek Kanhangad, 2019. PVSNet: Palm vein authentication siamese network trained using triplet loss and adaptive hard mining by learning enforced domain specific features. In *2019 IEEE 5th international conference on identity, security, and behavior analysis (ISBA)* (pp. 1-8). IEEE. DOI: https://doi.org/ 10.1109/ISBA.2019.8778623.D
11. Yandong Wen, Kaipeng Zhang, Zhifeng Li, and Yu Qiao, 2016. A discriminative feature learning approach for deep face recognition. In *European conference on computer vision* (pp. 499-515). Springer, Cham. DOI: https:// doi.org/10.1007/978-3-319-46478-7_31.
12. Weihua Chen, Xiaotang Chen, Jianguo Zhang, and Kaiqi Huang, 2017. Beyond triplet loss: a deep quadruplet network for person re-identification. In *Proceedings of the IEEE conference on computer vision and pattern recognition* (pp. 403-412). DOI: https://doi.org/ 10.1109/CVPR.2017.145.D
13. Jian Wang, Feng Zhou, Shilei Wen, Xiao Liu, and Yuanqing Lin, 2017. Deep metric learning with angular loss. In *Proceedings of the IEEE International Conference on Computer Vision* (pp. 2593-2601). DOI: https://doi.org/ 10.1109/ICCV.2017.283.
14. Weiyang Liu, Yandong Wen, Zhiding Yu, Ming Li, Bhiksha Raj, and Le Song, 2017. Sphereface: Deep hypersphere embedding for face recognition. In *Proceedings of the IEEE conference on computer vision and pattern recognition* (pp. 212-220). DOI: https://doi.org/10.1109/CVPR.2017.713.
15. Andrew G. Howard, Menglong Zhu, Bo Chen, Dmitry Kalenichenko, Weijun Wang, Tobias Weyand, Marco Andreetto, and Hartwig Adam, 2017. Mobilenets: Efficient convolutional neural networks for mobile vision applications. *arXiv preprint arXiv:1704.04861*. retrieved from https://arxiv.org/pdf/1704.04861.
16. Kaiming He, Xiangyu Zhang, Shaoqing Ren, and Jian Sun, 2015. Delving deep into rectifiers: Surpassing human-level performance on imagenet classification. In *Proceedings of the IEEE international conference on computer vision* (pp. 1026-1034). DOI: https://doi.org/ 10.1109/ICCV.2015.123.
17. Sergey Ioffe and Christian Szegedy, 2015. Batch normalization: Accelerating deep network training by reducing internal covariate shift. In *International conference on machine learning* (pp. 448-456). PMLR.

18. Jun Cheng, 2017. Brain Tumor Dataset. figshare. DOI: 10.6084/m9.figshare.1512427.v5.
19. Parnian Afshar, Konstantinos N. Plataniotis, and Arash Mohammadi, 2019. Capsule networks for brain tumor classification based on MRI images and coarse tumor boundaries. In *ICASSP 2019-2019 IEEE International Conference on Acoustics, Speech and Signal Processing (ICASSP)* (pp. 1368-1372). IEEE. DOI: https://doi.org/10.1109/ICASSP.2019.8683759.
20. Ruihang Chu, Yifan Sun, Yadong Li, Zheng Liu, Chi Zhang, and Yichen Wei, 2019. Vehicle re-identification with viewpoint-aware metric learning. In *Proceedings of the IEEE/CVF International Conference on Computer Vision* (pp. 8282-8291). DOI: https://doi.org/10.1109/ICCV.2019.00837.
21. Mesut Toğaçar, Burhan Ergen, and Zafer Cömert, 2021. Tumor type detection in brain MR images of the deep model developed using hypercolumn technique, attention modules, and residual blocks. *Medical & Biological Engineering & Computing*, *59*(1), 57-70. DOI: https:// doi.org/10.1007/s11517-020-02290-x.
22. Ashish Vaswani, Noam Shazeer, Niki Parmar, Jakob Uszkoreit, Llion Jones, Aidan N. Gomez, Łukasz Kaiser, and Illia Polosukhin, 2017. Attention is all you need. In *Advances in neural information processing systems* (pp. 5998-6008). DOI: https://doi.org/10.1145/1218913.1218915.
23. Abdu Gumaei, Mohammad Mehedi Hassan, Md Rafiul Hassan, Abdulhameed Alelaiwi, and Giancarlo Fortino, 2019. A hybrid feature extraction method with regularized extreme learning machine for brain tumor classification. *IEEE Access*, *7*, 36266-36273. DOI: https://doi.org/ 10.1109/ACCESS.2019.2904145.
24. Mustafa R. Ismael, and Abdel-Qader Ikhlas, 2018. Brain tumor classification via statistical features and back-propagation neural network. *IEEE international conference on electro/information technology (EIT)*, pp. 0252-0257. IEEE, 2018. DOI: 10.1109/EIT.2018.8500308
25. Taranjit Kaur, and Tapan Kumar Gandhi, 2020. Deep convolutional neural networks with transfer learning for automated brain image classification. *Machine Vision and Applications* 31, no. 3 (2020): 1-16. DOI: https://doi.org/10.1007/s00138-020-01069-2